\documentclass[letterpaper, 10 pt, conference]{ieeeconf} 

\IEEEoverridecommandlockouts                              

\usepackage{xcolor}
\usepackage[pdftex]{graphicx}
\graphicspath{{figures/}}
\usepackage{hyperref}

\usepackage[caption=false,font=footnotesize]{subfig}

\title{\LARGE \bf
Learning Early Social Maneuvers for Enhanced Social Navigation
}

\author{
    Yigit~YILDIRIM, Mehmet SUZER and~Emre~UGUR
    \thanks{Authors are with the Computer Engineering Department, Bogazici University. Corresponding email address: \texttt{yigit.yildirim@bogazici.edu.tr}}
}

\begin{document}

\maketitle
\thispagestyle{empty}
\pagestyle{empty}

\begin{abstract}

Socially compliant navigation is an integral part of safety features in Human-Robot Interaction. Traditional approaches to mobile navigation prioritize physical aspects, such as efficiency. Still, the social competence of robots has gained traction as the attributed trust towards them is a crucial factor influencing their acceptance in our daily lives. Recent techniques to improve the social compliance of navigation often rely on predefined features or reward functions, introducing assumptions about social human behavior. To address this limitation, we propose a novel Learning from Demonstration (LfD) framework for social navigation that exclusively utilizes raw sensory data to address this limitation. Additionally, the proposed system contains mechanisms to consider the future paths of the surrounding pedestrians, acknowledging the temporal aspect of the problem. The final product is expected to reduce the anxiety of people sharing their environment with a mobile robot, helping them trust that the robot is aware of their presence and will not harm them. As the framework is currently being developed, we outline its components, present experimental results, and discuss future work towards realizing this framework.

\end{abstract}

\section{Introduction}
\begin{figure}[t]
    \begin{center}
        \includegraphics[width=0.45\columnwidth]{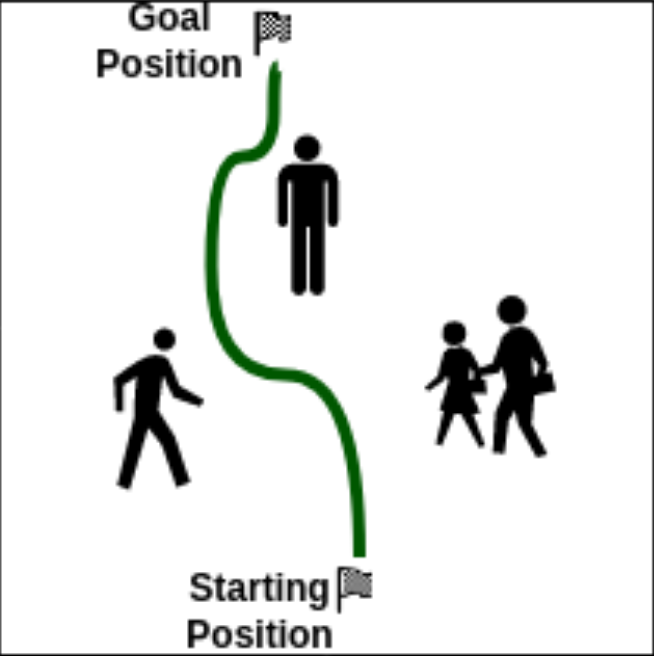}
        \caption{A socially compliant navigation path. Even though the path is suboptimal regarding time and energy, it is socially more acceptable.}
        \label{fig:sn}
    \end{center}
\end{figure}

It is common for many to see other people peeping at a robot with apprehension as it passes nearby. As supported by Human-Robot Interaction (HRI) studies (e.g. \cite{bartneck2020human}), the spatial positioning of the mobile robot is one of the contributors to the persistent hostile attitude towards them. However, conventional approaches to mobile robot navigation have addressed only the physical aspects of the problem: improving the efficiency \cite{rosmann2017integrated} and realizing collision-free navigation \cite{khatib1986real, burgard1999experiences}. As robots have started penetrating our daily environments, robotic systems are expected to exhibit social competence to improve the attributed trustworthiness, which facilitates their acceptance in social environments \cite{townsend2022trust}. From the navigation perspective, social competence might include avoiding personal zone intrusions, not passing between a group of people, etc. A complete list of scenarios is given in \cite{francis2023principles}. In broad terms, robot navigation that complies with the social norms of the people is called socially compliant or social navigation \cite{kretzschmar2016socially}. The sketch in Fig. \ref{fig:sn} describes this behavior. For a formal description, see \cite{mavrogiannis2023core}.

Researchers have adopted different strategies to increase the social compatibility of robotic navigation. While some studies take the indirect strategy of decreasing the discomfort \cite{diego2011please} by avoiding confrontations with people, the majority aim at increasing the comfort of the people around the robot. For that purpose, early studies used hand-crafted utility functions to steer away from the people around. Social Force Model (SFM), \cite{helbing1995social}, is one of the early models that adopted this idea. On the contrary, \cite{vasquez2014inverse} argues that social compliance requires more flexible policies, and data-driven techniques provide this capability. In this respect, Reinforcement Learning (RL) offers improvements, as shown in \cite{chen2017decentralized}, but the requirement for a carefully designated reward function leads to making unproven assumptions about socialness.

To alleviate the reward engineering requirement and assumptions about the human navigation model, frameworks that follow the Learning from Demonstration (LfD) paradigm offer a remedy. Inverse Reinforcement Learning (IRL) is a widely employed LfD framework for social navigation, \cite{abbeel2004apprenticeship}. IRL uses demonstration data to find the reward function, leading to the policy that the agent might be following. Even though IRL approaches learn the reward function from expert demonstrations, the need for predefined features transforms the assumptions made on the reward level to the feature level. A purely data-driven LfD technique that extracts features from the data would not need to rely on such suppositions. Furthermore, \cite{taniguchi2023world} describes the value of continuously modeling and predicting the world for autonomous agents. In \cite{zanlungo2011social}, the authors state their belief that the trajectory prediction of other pedestrians influences the navigational path of a person. More recent studies, e.g., \cite{karnan2022socially}, also share this perspective but propose different ways to address the issue, such as modeling navigational interactions with spatiotemporal representations. Therefore, a more holistic approach should inevitably consider the future positions of pedestrians to model and realize social navigation properly.

This paper presents our envisioned LfD framework to incorporate early maneuvers into the social navigation frameworks for improved social compliance, one of the safety features affecting HRI, \cite{zacharaki2020safety}. This novel framework will rely only on the raw sensory data, not on any external assumptions, as explained in Section \ref{sec:met}. Also, an integrated trajectory-forecasting mechanism is planned to address the temporal aspect of the navigation. After a literature review in Section \ref{sec:rel}, the individual constituents of the proposed framework are presented in Section \ref{sec:met}. Later, in Section \ref{sec:exp}, we present the latest findings with the current state of these modules. Finally, in Section \ref{sec:con}, we share our conclusions and propose a plan to implement the framework.

\section{Related Work}
\label{sec:rel}

One of the critical factors affecting the success of human-robot interaction is the comprehensibility of the robot's actions for humans \cite{kirtay2020modeling}. The navigation induces nonverbal communication, and social competence in this type of communication also influences the perceived trustworthiness of the robot \cite{singamaneni2024survey}. Earlier techniques to improve the social competence of robot navigation mostly use manually designed controllers \cite{zanlungo2011social}. These controllers move the robot according to predefined rules that increase the socialness of the navigation based on related assumptions. The proxemics theory described in \cite{hall1966hidden} provided a basis for subsequent studies by defining the abstract social zones around the people. Although the robot navigates toward a global target, these approaches are reactive in that they only consider the local surroundings of the robot, which often leads to the freezing robot problem \cite{trautman2015robot}.

\begin{figure}[t]
    \begin{center}
        \includegraphics[width=0.95\columnwidth]{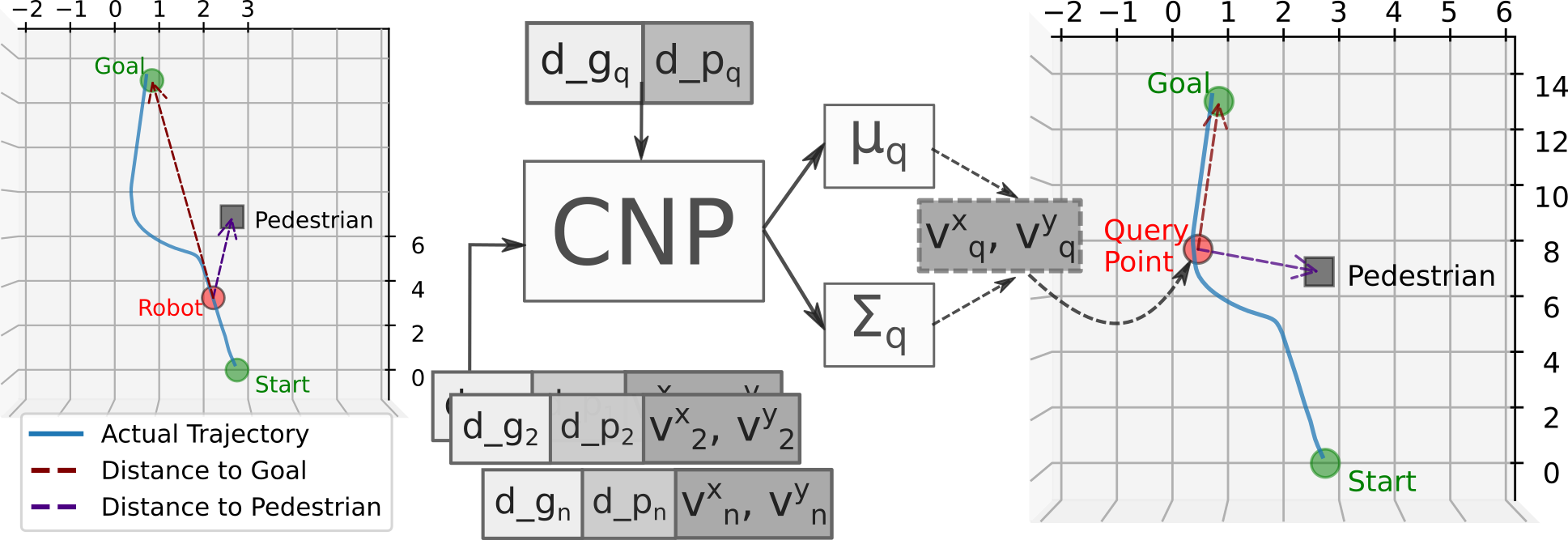}
        \caption{The local planner module used in \cite{yildirim2021learning}. In this model, the neural network is fed information about the robot's destination and the position of the closest pedestrian, and the appropriate velocity commands are output.}
        \label{fig:sn_local}
    \end{center}
\end{figure}

Due to the advances in data-driven control approaches, researchers have proposed more flexible frameworks. As new metrics have been proposed to assess the socialness of a navigation system, the use of RL is justified with a reward function that aims to maximize the performance of these metrics. Therefore, with carefully designated objective functions that describe the social context of the robot, RL-based approaches are employed in studies such as \cite{chen2017decentralized, chen2019crowd, long2018towards} to teach robots socially compliant navigation policies. On the other hand, to the best of our knowledge, there is no consensus on what social is and how it reflects on pedestrians yet in the literature. Without such a formal description, these reward functions and metrics are only partially sufficient to explain this social behavior.

To avoid (1) the reward-engineering requirement and (2) searching for a suboptimal policy because of the partially descriptive objective function, directly modeling and imitating human navigation is a reasonable approach. The LfD paradigm provides different means of achieving this goal. Conventionally, IRL is one of the most employed approaches in the social navigation domain. Researchers in earlier IRL applications, such as \cite{kretzschmar2016socially, vasquez2014inverse}, represent the reward function as a linear combination of predefined features. As expected, with more extensive feature sets, the resemblance of the extracted policy to actual human navigation increases. On the other hand, similar considerations about reward engineering also apply here. In \cite{wulfmeier2017large}, researchers apply Maximum Entropy Deep IRL (MEDIRL) to directly learn social compliance from the raw sensory data only in static environments as more complex mechanisms are needed to achieve the same goal in realistic environments.

Other LfD approaches, such as Imitation Learning and Behavior Cloning (BC), attempt to learn the policy directly, skipping the reward function. In \cite{karnan2022socially}, a BC agent is trained to realize socially compliant navigation using the data collected in real-world settings with a teleoperated robot. Using actual navigation data eliminates the criticism that simulation-based studies face: using only the simulated navigation data, data-driven models can, at best, learn the model that already controls the simulated agents. In \cite{yildirim2022learning}, researchers again used real-world data to improve the social navigation capabilities using a framework consisting of two Conditional Neural Processes (CNPs) \cite{garnelo2018conditional}. CNP is an LfD framework that contains an Encoder and a Query Network to model and synthesize skills demonstrated by experts. Accordingly, the mechanism given in Fig. \ref{fig:sn_local} is proposed as the local planner where a CNP-based module learns to control the robot, incorporating social maneuvers into kinematics calculations. Aside from the fact that joining these two different problems together unessentially complicates the learning process, this framework has another shortcoming: it neglects the temporal aspect of human navigation by assuming that only the k-closest pedestrians affect the navigation trajectory.

In this study, we introduce a novel social navigation framework that is currently under development and aims to improve the one proposed in \cite{yildirim2022learning}. The intended framework considers several critical issues, such as avoiding predetermined rules about human comfort/discomfort, learning from the raw sensory data collected in the real world, and considering the forecasted paths of pedestrians. As the framework is currently being developed, in this paper, we will limit our focus to the individual components but not the entire system.

\section{Method}
\label{sec:met}

Pedestrians avoid spatial discomfort during navigation by collaboratively exhibiting social maneuvers. In mobile navigation, this kind of behavior can be achieved at the local controller level as it does not directly influence the destination of the navigation. Therefore, we propose to employ a CNP-based LfD module which is used to teach the robot spatial social norms, such as maneuvering at a distance, not invading personal zones, etc., in combination with an off-the-shelf local planner, such as the Dynamic Window Approach (DWA) described in \cite{fox1997dynamic}. This way, the robot can handle the necessary kinematics to realize the learned social maneuvers. In the following, we elaborate on these improvements.

\subsection{Incorporating the Future Predictions}
\label{sec:fcsn}

\begin{figure}[t]
    \begin{center}
        \includegraphics[width=0.57\columnwidth]{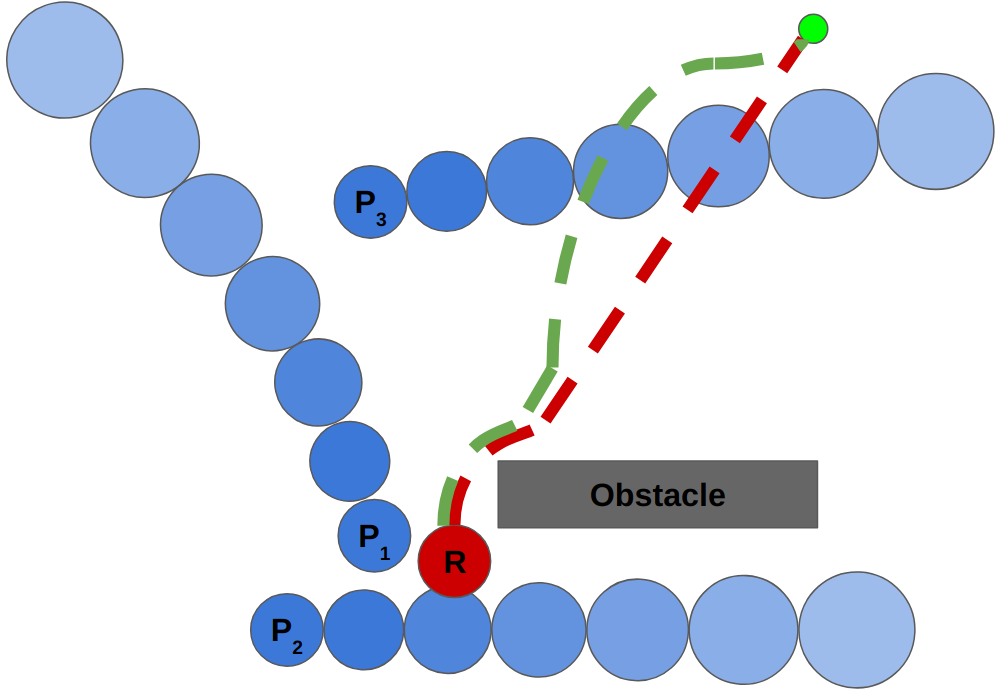}
        \caption{An environment with three pedestrians, a robot, and an obstacle. The nonsocial behavior is illustrated with the red path. Even though $P_1$ and $P_2$ are closer to the robot, the pedestrian whose trajectory affects the robot is $P_3$, since only $P_3$ presents the chance of encounter. Therefore, the robot should follow the green path to avoid future encounters.}
        \label{fig:fc}
    \end{center}
\end{figure}

In contrast to what is proposed in \cite{yildirim2022learning}, it is evident that the spatial proximity of others is not the sole determinant of an agent's navigation trajectory. People probably precompute other's trajectories, and based on this prediction, they decide on their path in advance, as in \cite{zanlungo2011social}, requiring to consider the trajectories not just spatially but also temporally. To depict this idea, a scenario is illustrated in Fig. \ref{fig:fc} where trajectory predictions of pedestrians are given with fading circles to indicate the increasing uncertainty. Due to the explicit social behavior in such scenarios, we believe that it is inevitable to consider future encounters in path planning.

\begin{figure}[b]
    \begin{center}
        \includegraphics[width=0.8\columnwidth]{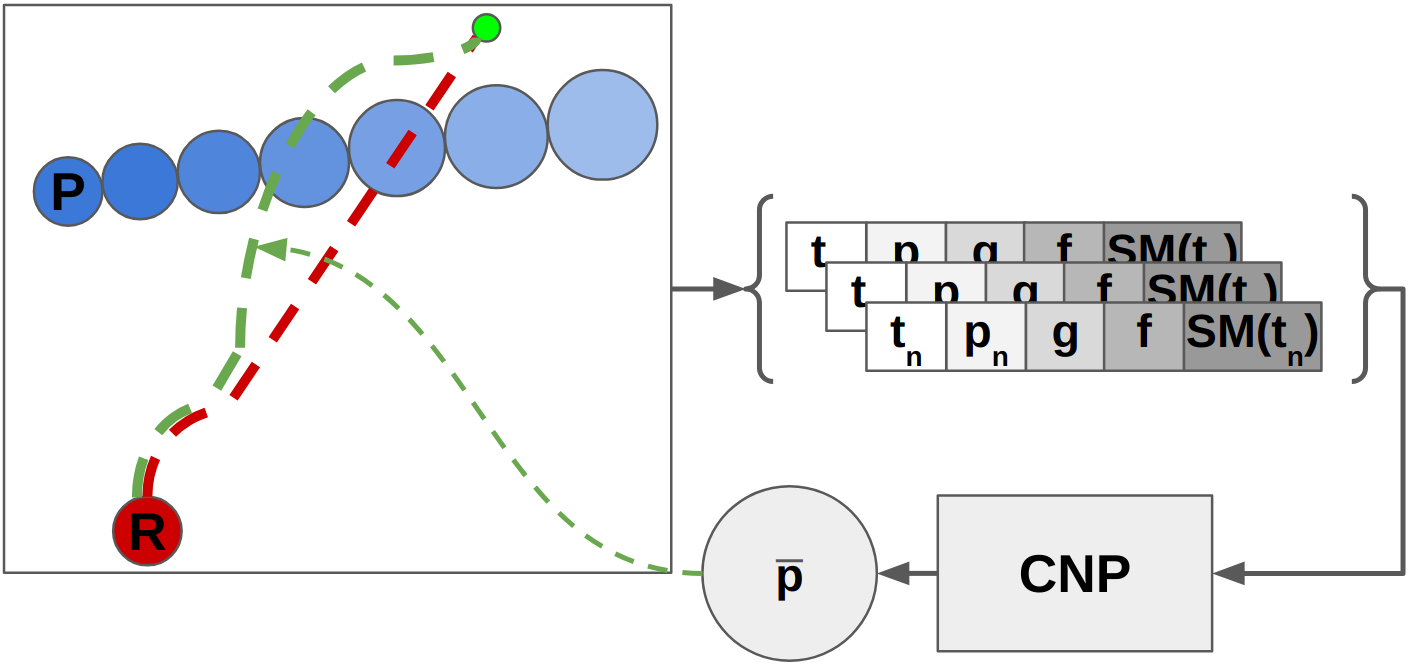}
        \caption{To include predicted future trajectories in the planning mechanism, the proposed approach is to feed them to the CNP-based LfD architecture. The neural network takes in tuples of the timestep, current position, navigation goal, the predicted trajectory of the pedestrian, and the robot's trajectory within the next few time steps. It outputs the predicted trajectory in a local window, allowing it to be converted into a local planner.}
        \label{fig:fccnmp}
    \end{center}
\end{figure}

To predict the trajectories of other agents, we use an LSTM-based RL approach, given in \cite{everett2018motion}. Despite the presumably incomplete objective that RL approaches have, we believe that this approach provides tolerable forecasts for the trajectories of others. To include this information in the robot's local planner, we propose the mechanism given in Fig. \ref{fig:fccnmp}. 

\subsection{Improving the Environmental Awareness}
\begin{figure}[t]
    \begin{center}
        \includegraphics[width=0.99\columnwidth]{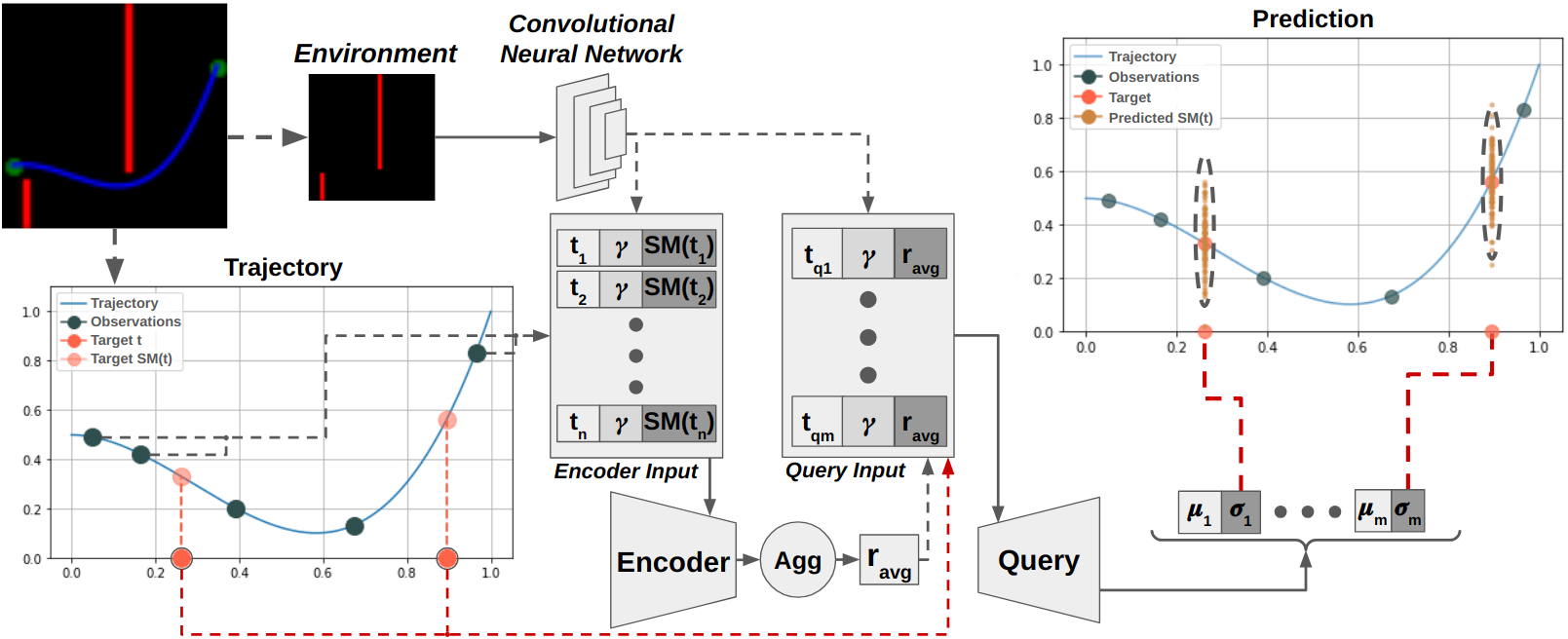}
        \caption{The use of a CNN as state encoder. The environment is presented as a 2D image where red bars signify the randomly placed obstacles, and the blue trajectory shows the expert demonstration. CNN processes the image and produces a fixed-size state encoding, which is concatenated to the Encoder's input. The Query produces predictions using this information.}
        \label{fig:conv-cnmp}
    \end{center}
\end{figure}

The assumption that only a fixed number of pedestrians influence a person's trajectory is incorrect but required for learning systems as the computation graph for the network is prepared before the training. To relax this constraint, following the idea presented in \cite{seker2019conditional}, we propose to use a Convolutional Neural Network (CNN) as a state encoder to incorporate the entire environment information into trajectory learning. The proposed model is given in Fig. \ref{fig:conv-cnmp}, where the environment is represented as an image. In the final version of the proposed framework, we aim to directly use the output of a $360^\circ$ lidar. The Encoder Network takes in $(t, \gamma$ and $\textit{SM}(t))$ tuples where t denotes the timestep, SM the sensorimotor response of the robot, and $\gamma$ contains the state encoding, which is produced by the CNN. Its output is concatenated with query timesteps to obtain the input for the Query Network. In the end, the Query Network outputs a mean ($\mu_q$) and a standard deviation ($\sigma_q$) for each query ($t_q$), producing SM predictions as normal distributions. The actual loss value is computed as the negative log-likelihood of the actual SM values under these predicted distributions. This function is given in Equation \ref{eq:loss}, where the \textit{softplus} is a utility function, ensuring the positivity of the standard deviation. The entire system is trained end-to-end with this loss.

\begin{equation}
\label{eq:loss}
\mathcal{L} = -\: log\: P(\:\textit{SM}(t_q)\: |\: \mathcal{N}(\mu_q,\: softplus(\sigma_q)\:)\:).
\end{equation}

Utilizing a $360^\circ$ lidar and the CNN is expected to eliminate the assumption that only the k-closest pedestrians influence the social trajectory of the robot.

\section{Experiments and Results}
\label{sec:exp}

\subsection{The Incorporation of Future Trajectories}
\begin{figure}[t]
    \centering
    \subfloat[Predicted trajectory of a robot (shown with \textcolor{red}{$\bullet$}) in the presence of a stationary pedestrian (shown with \textcolor{blue}{\textbf{x}}.)]{\label{fig:snWst}\includegraphics[width=0.47\columnwidth]{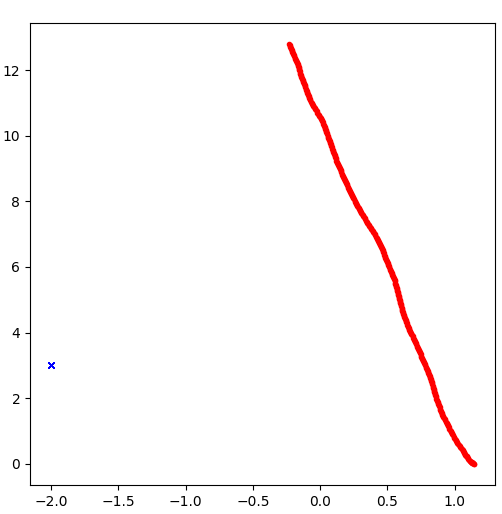}}
    \enspace
    \subfloat[Predicted trajectory of the robot where the other pedestrian moves indifferently towards a destination.]{\label{fig:4_snWfc}\includegraphics[width=0.47\columnwidth]{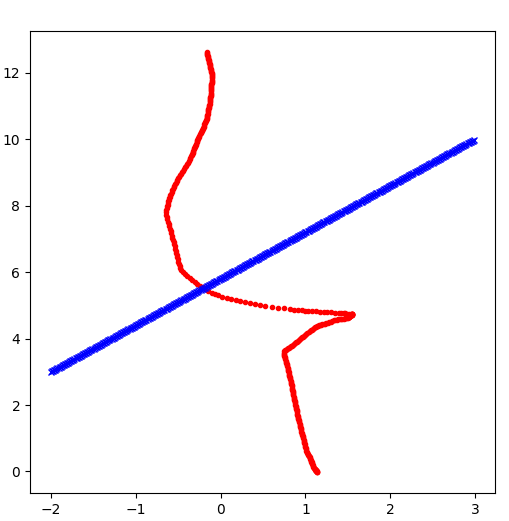}}
    \caption{Effects of the predicted trajectories on the planned trajectory of the robot. In either case, the robot's motion starts and ends at the same location.}
    \label{fig:fc_res}
\end{figure}

As noted earlier, we have not yet integrated these components into a working framework. Therefore, current results fail to represent the collaborative nature of this phenomenon. On the other hand, under the assumption that other components are working as expected, the influence of others can be modeled by the approach given in Section \ref{sec:fcsn}. Given that the trajectory forecasting mechanism produces predictions for pedestrians, our CNP-based model learns better trajectories. Two scenarios are presented in Fig. \ref{fig:fc_res}.

From these results, it is apparent that this component can effectively incorporate future predictions into the robot's local planner. In combination with the CNN, a changing number of pedestrian trajectories can be modeled in our system.

\subsection{The use of CNN as a state encoder}
Currently, we do not have the results with the data collected in real-world settings. Nonetheless, we provide the results of the experiments on synthetically generated environments. These environments consist of 2 randomly placed obstacles of random heights and random start and destination positions. The trajectory is found using an obstacle-avoiding A* search and smoothed using polynomial regression.

After the generation of these environments, they are used to train the model. The environment without the trajectory is given to the CNN to create state encodings, $\gamma$, which is, in turn,  concatenated with the samples of the trajectory-learning CNP. CNP produces target SM values that produce a loss, which is backpropagated to adjust network parameters. All neural networks; CNN, Encoder, and Query, are trained end-to-end with the same loss.

\begin{figure}[t]
    \begin{center}
        \subfloat{\includegraphics[width=0.44\columnwidth]{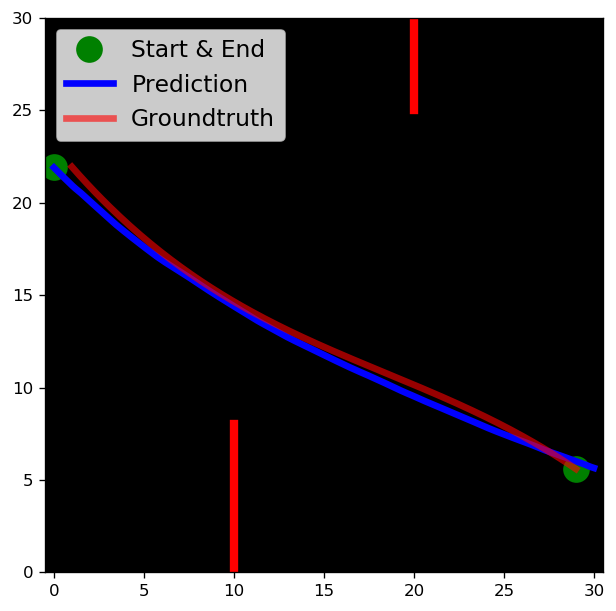}}
        \enspace
        \subfloat{\includegraphics[width=0.44\columnwidth]{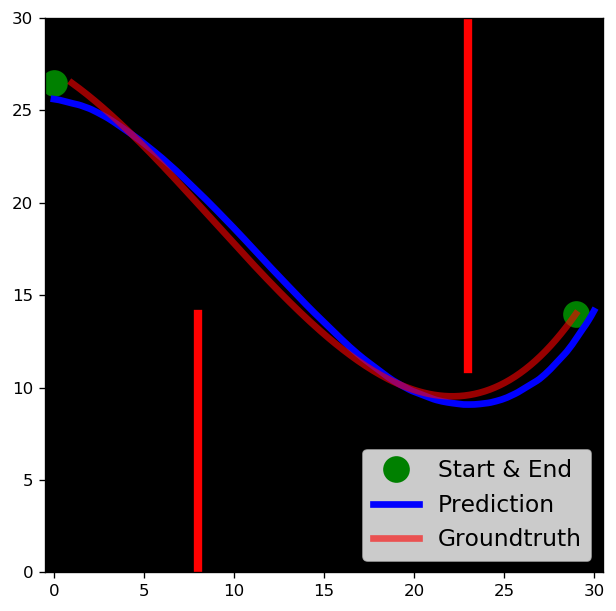}}\\
        \subfloat{\includegraphics[width=0.44\columnwidth]{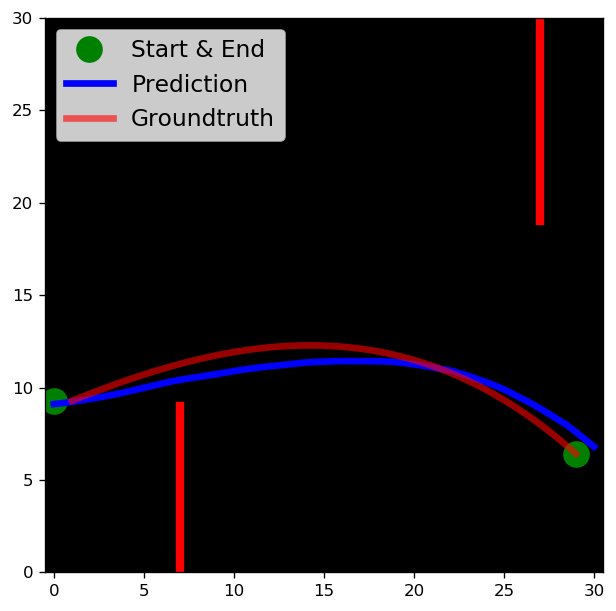}}
        \enspace
        \subfloat{\includegraphics[width=0.44\columnwidth]{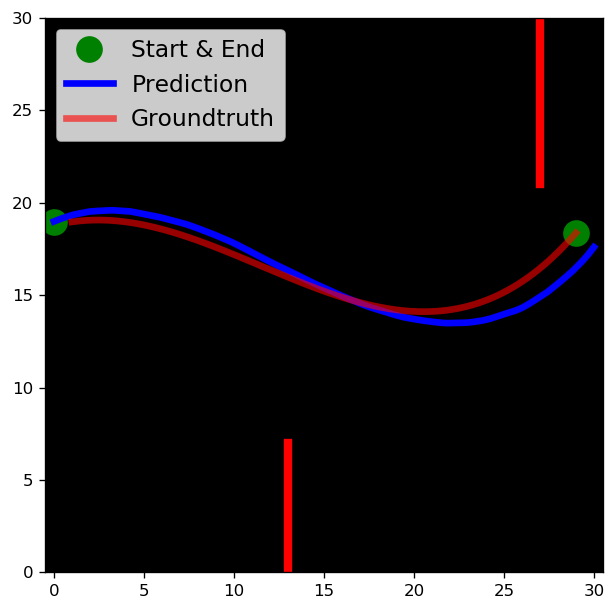}}
        \caption{Example results of the utilization of CNN use as a state encoder in combination with the CNP. Here, a CNN is fed with the environmental information, and a CNP is tasked to learn the trajectory that avoids obstacles.}
        \label{fig:res_cnn}
    \end{center}
\end{figure}

The results presented in Fig. \ref{fig:res_cnn} show that it is possible to encode high-dimensional state information with CNNs to use them later to model trajectories. When used instead of the image of the environment, we expect to see a similar performance with the point cloud data that the lidar sensor produces.

\section{Conclusion and Future Work}
\label{sec:con}

Our exploration of Learning from Demonstration (LfD) frameworks for social navigation highlights the necessity for methods that alleviate the reliance on predefined features or reward functions. While existing approaches, such as Inverse Reinforcement Learning (IRL), offer promising avenues, they still impose assumptions about human behavior at either the reward or feature level. Our proposed framework aims to overcome these limitations by leveraging raw sensory data exclusively, paving the way for a more adaptable and naturalistic approach to social navigation.

This position paper presents the state of the individual components of the envisioned navigation system. Even though the individual components perform as expected, it is not possible to infer the final performance of the entire system. Moving forward, our focus will be shifted towards evaluating the abovementioned components with real-world data. Although data resources are scarce in social navigation, two recent studies, \cite{karnan2022socially} and \cite{nguyenmusohu}, have significantly contributed by providing datasets collected in the real world. If the components we propose succeed on these actual data, we plan to combine them. The trajectory-forecasting module is planned to make predictions for pedestrians in a local window around the robot. The size of this window will be decided upon initial evaluations of the entire system since the size and appearance can contribute to the robot's perceived trustworthiness \cite{zlotowski2016appearance}. The forecasted trajectories will be encoded with a CNN, and the trajectories are planned to be learned by the CNP-based module. Additionally, we plan to investigate the scalability of our framework and its practical applicability to mobile robots. To achieve this, we plan to integrate the entire framework into an off-the-shelf local planner, such as DWA, to be easily deployed on actual mobile robots.

Subsequently, we want to evaluate our claim of enhanced social compliance and trust in robot navigation. This involves (1) conducting experiments in various real-world scenarios to assess the overall performance of the entire system and (2) evaluating the psychological effects introduced by the interaction with the robot. The scenarios and metrics used in the quantitative evaluation will be taken from \cite{francis2023principles} while ideas presented in \cite{kirtay2021learning} will be followed for the psychological assessment of this nonverbal interaction. 

Currently, unimodal trajectory forecasts are being used in the system. In the future, this study can be expanded to pay appropriate attention to the multimodal nature of navigation. To this end, utilization of intention estimation mechanisms, such as \cite{amirshirzad2022adaptive}, and multimodal LfD frameworks, such as \cite{yildirim2024conditional}, should be considered. Ultimately, we aim to develop a data-driven solution that enables robots to navigate socially compliantly, fostering greater acceptance and integration into human-centric settings.

\section*{Acknowledgement}
This research is supported by the European Union under the INVERSE project (101136067). The source codes are available at \url{https://github.com/yildirimyigit/sn\_w\_cnps}.

\bibliographystyle{IEEEtran}
\bibliography{ref}

\end{document}